\documentclass[a4paper]{llncs}

\usepackage[T1]{fontenc}
\usepackage{graphicx}
\usepackage{booktabs}
\usepackage{amsmath}
\usepackage{amssymb}
\usepackage{xspace}
\usepackage{array}
\usepackage{tabularx}
\usepackage{microtype}
\usepackage{url}

\ifdefined\pdfpagewidth
\fi

\newcommand{\method}{AITHER\xspace}
\newcommand{\sfr}{Semantic-Frame Router\xspace}
\newcommand{\framebench}{FRAME\xspace}
\newcommand{\sicans}{A_{\mathrm{sic}}}
\newcommand{\frameans}{A_{\mathrm{frame}}}
\newcommand{\imgans}{A_{\mathrm{img}}}
\newcommand{\txtans}{A_{\mathrm{txt}}}
\newcommand{\finalans}{A_{\mathrm{final}}}
\newcommand{\fallbackans}{A_{\mathrm{fb}}}
\newcommand{\laaans}{A_{\mathrm{laa}}}
\title{AITHER: Authority-Informed Evidence Routing for Mixed-Authority Vision-Language Reasoning}
\author{Xinjin Li\inst{1}\textsuperscript{*} \and
Haoyue Zhang\inst{2}\textsuperscript{*} \and
Xinyi Gou\inst{3}\textsuperscript{*} \and
Yuruo Gong\inst{4} \and
Yuanjian He\inst{5} \and
Muzi Li\inst{6} \and
Yujian Long\inst{7} \and
Yu Ma\inst{8} \and
Yiliu Xu\inst{8} \and
Tianxin Zhou\inst{9}\textsuperscript{\ensuremath{\dagger}}}
\institute{Columbia University, United States of America
\and UC Berkeley, United States of America
\and Texas A\&M University, United States of America
\and Rice University, United States of America
\and Sun Yat-sen University, China
\and Stanford University, United States of America
\and Georgetown University, United States of America
\and Carnegie Mellon University, United States of America
\and University of Southern California, United States of America\\
\textsuperscript{*} These authors contributed equally.\\
\textsuperscript{\ensuremath{\dagger}} Corresponding author: \email{zhoutx0@gmail.com}}

\begin{document}
\pagestyle{empty}
\maketitle

\begin{abstract}
Reliable vision-language reasoning requires a control decision that conventional grounding methods leave implicit: which input is entitled to determine the answer? Auxiliary text may be a defeasible claim about an image, a binding task instruction, or compatible context. Correcting every image--text conflict toward the image therefore repairs false claims but overwrites valid task rules. We formalize this setting as \emph{mixed-authority vision-language reasoning} and introduce \method, an \emph{Authority-Informed Text--Image Hierarchy for Evidence Routing}. Its final policy, the \sfr, is a training-free black-box control layer that (i) compiles explicit task directives, (ii) recovers standardized spatial frames from metadata-scrubbed pixels, (iii) resolves the semantic polarity of remaining notes, and (iv) routes among frozen observable candidates with fail-closed semantics. This ordered decomposition converts authority arbitration from an unconstrained prompt preference into a reproducible inference program. Across 1,620 MATRIX predictions, \method reaches 95.1\% accuracy, exceeding the strongest task-and-frame baseline by 9.1 points with a 95\% record-cluster interval of 6.8--11.5 points and no harmed records. It further reaches 100.0\% on 1,800 controlled \framebench predictions and 93.3\% on GQA-MATRIX. The results show that explicit authority routing can simultaneously improve false-claim correction, preserve task-defined answers, and eliminate clean-context overcorrection.
\end{abstract}

\section{Introduction}

Large vision-language models jointly consume images, questions, task instructions, and auxiliary context~\cite{li2023blip2,liu2023llava}. Their reliability is usually framed as a grounding problem: when language conflicts with pixels, the system should recover the visually supported answer. This formulation is necessary but incomplete. The same textual channel can contain a claim about the scene, a rule that defines the required response, or context that already agrees with the image. These inputs may exhibit similar surface disagreement while carrying fundamentally different authority over the output.

Consider an image that supports option A while an accompanying note mentions option B. If the note is a caption-like claim, the system should reject it and follow the image. If the note instead states ``for this task, return option B,'' it is not evidence to be verified; it is the specification of the desired answer. A conflict detector, visual-first prompt, or unconditional image rollback observes the same disagreement in both cases and therefore cannot choose the correct action. The decisive question is not only whether the sources disagree, but \emph{which source is authorized to decide}.

Prior work has exposed language priors in visual question answering~\cite{goyal2017vqa2,agrawal2018dont,lee2025vlind}, benchmarked image-unsupported generations~\cite{li2023pope,wang2023amber,guan2024hallusionbench,cao2024visdiahalbench}, and analyzed model-dependent modality preference under conflicting inputs~\cite{pezeshkpour2025mixed}. Hallucination mitigation verifies or revises claims~\cite{yin2023woodpecker,zhou2024lure}, modifies decoding~\cite{huang2024opera,leng2024vcd}, or searches for stronger visual prompts~\cite{woo2025bbvpe}. These methods strengthen visual faithfulness, yet visual faithfulness is not the correct objective when text legitimately defines the task. Instruction-hierarchy research supplies the complementary concept of privileged instructions~\cite{wallace2024instructionhierarchy}, while black-box verification shows that observable intermediate decisions can support structured correction~\cite{dhuliawala2024chain}. What remains missing is a multimodal control layer that resolves authority \emph{before} correction.

We formulate this missing capability as \emph{mixed-authority vision-language reasoning}. Each instance combines an image, a visual question, answer options, and an auxiliary note whose role is a defeasible visual claim, an authoritative task rule, or compatible context. The objective is to identify the governing source and then select an answer that satisfies three coupled requirements: correct false claims, preserve task-defined targets, and avoid perturbing compatible context. This formulation separates \emph{source conflict} from \emph{source authority} and exposes failure modes that are hidden by aggregate visual-consistency scores.

We introduce \method, an \emph{Authority-Informed Text--Image Hierarchy for Evidence Routing}. \method treats multimodal reliability as a control-plane problem over observable candidate answers. Its final executable policy is the \sfr: an ordered, deterministic program that first compiles binding task rules, then recovers standardized option frames from scrubbed pixels, then resolves the semantic polarity of unresolved notes, and finally chooses an image anchor, text probe, or SIC-preserving fallback. Each route is terminal once a higher-authority decision is available. Consequently, a plausible visual answer cannot overwrite a uniquely compiled task instruction, while an uncertain note cannot trigger speculative correction. The policy requires no access to model weights, logits, hidden states, provider identity, regime labels, or gold answers.

We evaluate \method in three complementary settings. MATRIX measures mixed authority on real images across existence, attribute, and relation questions. \framebench (\emph{Frame-Resolved Authority in Multimodal Evaluation}) isolates task and spatial precedence in controlled option-framed scenes. GQA-MATRIX provides a separate real-image transfer check. The evaluation covers 3,510 endpoint--item predictions and six heterogeneous black-box endpoints on the two principal benchmarks.

Our contributions are fourfold:
\begin{itemize}
    \item We identify \emph{authority} as a first-class variable in multimodal inference and formalize mixed-authority reasoning through distinct correction, preservation, and non-interference objectives.
    \item We introduce \method and its \sfr, a training-free black-box hierarchy that integrates instruction compilation, metadata-scrubbed frame recovery, semantic note resolution, observable candidate routing, and fail-closed fallback in one executable control program.
    \item We develop MATRIX and \framebench, complementary diagnostics that expose false-claim susceptibility, instruction overwriting, context overcorrection, and spatial-routing errors under balanced authority regimes.
    \item We establish the strongest performance among all evaluated methods: a 9.1-point MATRIX improvement over the task-and-frame hybrid with zero harmed records, a controlled 100.0\% result on \framebench, 93.3\% on GQA-MATRIX, and gains on all six principal endpoints.
\end{itemize}

\section{Related Work}

\subsection{Visual Grounding and Cross-Modal Conflict}
Visual question answering benchmarks have long shown that models can exploit linguistic regularities instead of image evidence~\cite{goyal2017vqa2,agrawal2018dont}. GQA extends this diagnosis to compositional and relational reasoning in real scenes~\cite{hudson2019gqa}, while VLind-Bench isolates language-prior blindness after controlling for prerequisite visual and commonsense capabilities~\cite{lee2025vlind}. Mixed Signals directly evaluates mismatched image--text inputs and finds that modality preference varies substantially with the model and task~\cite{pezeshkpour2025mixed}. These studies reveal when modalities disagree; our focus is the subsequent authority decision that determines which source may control the output.

\subsection{Multimodal Hallucination Evaluation and Mitigation}
POPE evaluates object hallucination through polling-based questions~\cite{li2023pope}. AMBER expands discriminative evaluation to existence, attribute, and relation hallucinations~\cite{wang2023amber}; HallusionBench combines visual illusion with language hallucination~\cite{guan2024hallusionbench}; and VisDiaHalBench measures the effect of misleading dialogue history~\cite{cao2024visdiahalbench}. These benchmarks motivate the false-claim regime of MATRIX, but do not distinguish a misleading claim from a legitimate instruction that intentionally changes the target.

Mitigation methods strengthen image consistency through post-hoc validation and revision~\cite{yin2023woodpecker,zhou2024lure}, model-internal decoding changes~\cite{huang2024opera,leng2024vcd}, or black-box visual-prompt selection~\cite{woo2025bbvpe}. Their common objective is to suppress content unsupported by the image. \method is orthogonal: it decides whether visual correction is authorized and then routes among already observable candidate answers.

\subsection{Instruction Authority and Black-Box Control}
The instruction hierarchy formalizes conflicts among textual instructions with different privilege levels~\cite{wallace2024instructionhierarchy}. Chain-of-Verification demonstrates that black-box generation can be decomposed into observable drafting and verification stages~\cite{dhuliawala2024chain}. \method connects these ideas to multimodal grounding by extending authority beyond text-only privilege: task rules, spatial cues, visual evidence, and compatible textual context are placed in a single precedence hierarchy. Unlike a learned role classifier, the final router is an auditable program whose route and fallback can be inspected for every prediction.

\section{Mixed-Authority Vision-Language Reasoning}

\subsection{Problem Formulation}
Each example is a tuple $x=(I,Q,T,O)$ containing an image $I$, visual question $Q$, auxiliary note $T$, and closed answer set $O$. Let $v(I,Q,O)$ denote the answer supported by the image and let $g(Q,T,O)$ denote the answer uniquely compiled from an explicit task rule. The semantic role of the note is
\begin{equation}
 r(Q,T)\in\{\mathrm{claim},\mathrm{rule},\mathrm{context}\}.
\end{equation}
The role-conditioned target is
\begin{equation}
 a^*(x)=
 \begin{cases}
 v(I,Q,O), & r(Q,T)=\mathrm{claim},\\
 g(Q,T,O), & r(Q,T)=\mathrm{rule},\\
 v(I,Q,O), & r(Q,T)=\mathrm{context}.
 \end{cases}
 \label{eq:target}
\end{equation}
A misleading claim contradicts visual evidence without changing the target. A task rule may contradict the image and intentionally redefine the target. Compatible context agrees with the visual answer and should remain undisturbed. Image--text disagreement is therefore an insufficient routing statistic: the correct decision is role-conditioned authority.

\subsection{Authority Objectives}
The formulation yields three objectives that conventional conflict accuracy conflates. \emph{Claim correction} requires routing away from a misleading textual claim toward a visual candidate. \emph{Task preservation} requires a uniquely compiled directive to remain invariant under competing visual evidence. \emph{Context non-interference} requires compatible text not to induce an unnecessary answer change. For a routing policy $\pi$, these requirements can be written as
\begin{align}
\mathcal{P}_{\mathrm{task}} &: \ g(Q,T,O)\in O \Rightarrow \pi(x)=g(Q,T,O),\\
\mathcal{C}_{\mathrm{claim}} &: \ r=\mathrm{claim} \Rightarrow \pi(x)\approx v(I,Q,O),\\
\mathcal{N}_{\mathrm{context}} &: \ r=\mathrm{context} \Rightarrow \pi(x)=v(I,Q,O).
\end{align}
The first property is enforceable by control flow; the latter two depend on the quality of observable candidates. This distinction motivates a hierarchy in which task authority is resolved deterministically, while evidence-dependent correction is opened only for the appropriate note role.

\section{AITHER: Hierarchical Semantic-Frame Routing}

\subsection{Architecture Overview}
\begin{figure}[t]
\centering
\includegraphics[width=\textwidth]{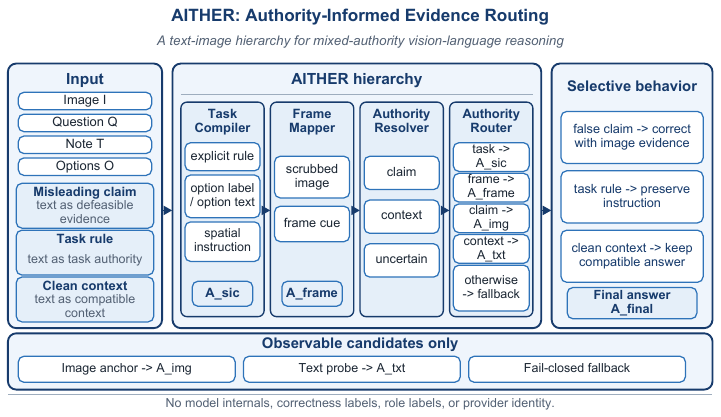}
\caption{Overview of \method. Binding task rules are resolved before spatial frame cues; unresolved notes are then assigned claim, context, or uncertain authority and routed to observable candidates. The hierarchy preserves valid instructions while selectively correcting false claims.}
\label{fig:overview}
\end{figure}

\method converts mixed-authority inference into an explicit control program. As shown in Figure~\ref{fig:overview}, the architecture contains four stages: a task compiler for binding directives, a frame mapper for standardized spatial cues, a semantic authority resolver for unresolved notes, and an ordered selector over frozen candidates. The paper-facing system is \method; the executable final policy in the accompanying implementation is the \sfr. Its central design principle is \emph{precedence with guarded route opening}: a downstream route is reachable only when every higher-authority stage is unavailable.

\subsection{Observable Candidate Interface}
For every frozen endpoint--item record, the runner exposes four parsed candidates:
\begin{align}
\sicans &\leftarrow \mathrm{SIC}(Q,T,O),\label{eq:sic-cand}\\
\frameans &\leftarrow \mathrm{FrameMap}(\mathrm{scrub}(I),O),\\
\imgans &\leftarrow \mathrm{CachedAnchor}(I,Q,O),\\
\txtans &\leftarrow \mathrm{CachedTextProbe}(Q,T,O).
\end{align}
The shared multimodal answer used by the baseline pipeline is denoted $\fallbackans$. A frozen legacy lexical action first produces $\laaans$: it uses $\fallbackans$ by default and retains the prior anchor/exclusion behavior on recognized claim templates. SIC replaces $\laaans$ only when an explicit instruction resolves to an answer target; hence $\sicans$ is either a compiled task answer or the unchanged legacy action. The image anchor and text probe are provider outputs generated before the final selector is executed. This interface isolates routing from candidate generation: \method changes no model parameter and observes no correctness field. It uses only the question and note text, answer-option text, metadata-scrubbed image pixels, and already saved candidate answers.

\subsection{Stage I: Spatial Instruction Compilation}
SIC converts explicit task language into an answer-space action rather than asking a model to reinterpret the instruction. It recognizes a fixed directive vocabulary (\emph{select}, \emph{choose}, \emph{pick}, \emph{submit}, and \emph{return}) and searches for one of three targets: an option label, matching option text, or named spatial position. Labels are validated against $O$; option strings are scanned in the frozen option order; and a spatial position is accepted only when it maps to exactly one option. Formally,
\begin{equation}
\mathrm{SIC}(Q,T,O,\laaans)=
\begin{cases}
 o, & o\in O\ \text{is resolved from an explicit directive},\\
 \laaans, & \text{otherwise}.
\end{cases}
\end{equation}
The resolution checks are deliberately conservative: an invalid label, unmatched text target, or non-unique spatial mapping does not create a high-authority route. Once a valid answer is compiled, selection terminates. This terminal branch enforces task preservation by construction and prevents any visual candidate from overwriting an explicit task specification.

\subsection{Stage II: Frame Recovery from Scrubbed Pixels}
When no task compiles, the router checks whether the image contains a standardized option frame. Frame recovery runs on an in-memory RGB copy in which the lower 12\% footer band is masked. The source image is never modified. For each position $p\in\{\mathrm{TL},\mathrm{TR},\mathrm{BL},\mathrm{BR}\}$, the detector evaluates a narrow rectangular boundary $\partial R_p$ around a scaled reference center and computes
\begin{equation}
 s_p(I)=\sum_{(u,v)\in\partial R_p}
 \mathbb{1}\!\left[\max_{c\in\{R,G,B\}} I_c(u,v)\le 60\right].
 \label{eq:frame-score}
\end{equation}
Let $p^*=\arg\max_p s_p$ and $s_{(2)}$ be the second-highest score. A frame is accepted only if the image is approximately square with side length 500--620 pixels, $s_{p^*}\ge\tau_s=400$, and $s_{p^*}-s_{(2)}\ge\tau_m=120$. The position is then mapped to an answer only when exactly one option string contains the corresponding spatial name. Otherwise the frame route remains closed. This stage therefore extracts the routing cue from pixels rather than reading benchmark metadata.

\subsection{Stage III: Semantic Authority Resolution}
If neither a task nor a frame determines the answer, the router resolves the semantic polarity of the note. A frozen lexical resolver checks claim patterns before context patterns and returns
\begin{equation}
 \rho(Q,T)\in\{\mathrm{claim},\mathrm{context},\mathrm{uncertain}\}.
\end{equation}
Claim-first precedence is conservative: a note that explicitly presents itself as a visual claim is treated as defeasible evidence even if it also contains generic agreement language. Context patterns open the text-probe route only when no claim cue is present. The resolver is deliberately deterministic rather than learned. Its purpose is not unrestricted discourse understanding, but a transparent authority decision over the templated note families evaluated here.

\subsection{Ordered Selection and Fail-Closed Semantics}
The final policy is
\begin{equation}
\finalans=
\begin{cases}
\sicans, & \text{if an explicit task target is resolved},\\
\frameans, & \text{if no task compiles and a frame maps to }O,\\
\imgans, & \text{if }\rho=\mathrm{claim}\text{ and }\imgans\in O,\\
\txtans, & \text{if }\rho=\mathrm{context}\text{ and }\txtans\in O,\\
\sicans, & \text{otherwise}.
\end{cases}
\label{eq:router}
\end{equation}

Equation~\ref{eq:router} encodes four design invariants. First, task compilation is terminal, yielding non-overwrite of valid instructions. Second, frame recovery precedes semantic note routing, so standardized spatial authority is read from image pixels. Third, claim and context routes select complementary candidates rather than applying one universal modality preference. Fourth, missing, ambiguous, or unparsed candidates return to SIC; the policy never invents an answer. The route is deterministic for fixed inputs and has linear text-processing cost plus a constant-size boundary scan for the four frame positions. Thus, the method adds a lightweight control layer without retraining or modifying the underlying endpoints.

\section{Experiments}

\subsection{Experimental Setup}
\paragraph{Benchmarks.}
MATRIX contains 270 real-image records derived from AMBER discriminative questions. It crosses three authority regimes---misleading claim, task rule, and clean context---with existence, attribute, and relation questions. Six endpoint responses per record produce 1,620 predictions, with 540 predictions per regime and per question type. \framebench contains 300 controlled option-framed records and six endpoint responses per record. Its 1,800 predictions are balanced across the three regimes, and each of the four answer labels appears 450 times. GQA-MATRIX contains 30 real-image records and three endpoint responses per record, adding 90 cross-dataset predictions.

\paragraph{Endpoints and replay protocol.}
The principal cache contains Gemma 3, GPT-4o-mini, Llama 3.2 Vision, MiMo-V2-Omni, Mistral Small 3.2, and Qwen3-VL-Plus. GQA-MATRIX contains Llama 3.2 Vision, MiMo-V2-Omni, and Qwen3-VL-Plus. We execute the final selector over saved provider outputs, so candidate generation is fixed and the evaluation isolates authority routing. Table~\ref{tab:config} summarizes the frozen protocol.

\begin{table}[t]
\caption{Core configuration of the final \sfr replay.}
\label{tab:config}
\centering
\small
\setlength{\tabcolsep}{4pt}
\begin{tabularx}{\textwidth}{lX}
\toprule
Component & Frozen setting\\
\midrule
Policy inputs & Question/note text, answer-option text, scrubbed image pixels, saved answer candidates\\
Task compiler & Valid label, matching option text, or uniquely mapped position under one of five directive verbs\\
Frame detector & Footer mask 12\%; dark threshold 60; score $\ge400$; top-two margin $\ge120$\\
Semantic resolver & Fixed claim-first, context-second lexical patterns; otherwise uncertain\\
Candidate source & Image anchor for claims; text probe for context; SIC for missing/uncertain routes\\
Split and scoring & Record-disjoint 80/20 split; exact parsed-option accuracy; record-level paired inference\\
Execution & Cache-only CPU replay; no provider calls, model download, training, or GPU\\
\bottomrule
\end{tabularx}
\end{table}

\paragraph{Baselines.}
Base and shifted chain-of-thought treat the note as ordinary context. Visual-first prompting explicitly prioritizes image evidence. Authority-aware prompting asks a model to resolve the role in one call. Legacy conflict rollback selects visual candidates whenever conflict is detected. A role-gated selector uses a provider-generated role probe. SIC compiles explicit directives, while SIC+FrameMap adds deterministic frame recovery. These baselines separate prompt bias, unconditional visual correction, provider-dependent role judgments, task compilation, spatial routing, and final semantic routing.

\paragraph{Metrics and statistical unit.}
Accuracy is the fraction of parsed answers equal to the gold option. Because the three regimes are balanced, overall accuracy is both the micro-average and regime macro-average. Endpoint outputs for the same record share an image, question, note template, and target; we therefore use the record, rather than the endpoint row, as the independent unit. Paired intervals are obtained by record-cluster bootstrap, and nonzero record differences are evaluated with an exact two-sided sign test.

\subsection{Stage-Wise Contribution of the Routing Hierarchy}
\begin{table}[t]
\caption{Stage-wise accuracy (\%) as deterministic components are added. The increments are complementary: task compilation dominates rule rows, frame mapping repairs controlled spatial cases, and semantic routing improves claim/context rows.}
\label{tab:stagewise}
\centering
\scriptsize
\setlength{\tabcolsep}{3.5pt}
\begin{tabular}{lrrr}
\toprule
Policy stage & MATRIX & \framebench & GQA-MATRIX\\
\midrule
Role-gated selector & 74.9 & 91.2 & 66.7\\
+ task compilation (SIC) & 86.0 & 99.1 & 74.4\\
+ frame mapping & 86.0 & 100.0 & 74.4\\
+ semantic routing (\method) & \textbf{95.1} & \textbf{100.0} & \textbf{93.3}\\
\bottomrule
\end{tabular}
\end{table}

Table~\ref{tab:stagewise} isolates the contribution of the hierarchy. SIC produces the largest gain on rule-defined examples by replacing a provider-dependent role judgment with a compiled task action. Frame mapping contributes on the controlled suite, where 16 SIC errors are repaired. The final semantic stage is decisive on natural images: it adds 9.1 points on MATRIX and 18.9 points on GQA-MATRIX by routing claims to image anchors and compatible context to text probes. The result is not a monolithic prompt improvement, but a sequence of non-substitutable control decisions.

\subsection{MATRIX: Main Mixed-Authority Results}
\begin{table}[t]
\caption{MATRIX accuracy over 1,620 endpoint--item predictions. Values are percentages; bold marks the best result.}
\label{tab:matrix-main}
\centering
\small
\setlength{\tabcolsep}{4pt}
\begin{tabular}{lrrrr}
\toprule
Method & Overall & Mislead & Task & Clean\\
\midrule
Shifted CoT & 69.2 & 44.8 & 66.7 & 96.1\\
Visual-first & 58.9 & 79.8 & 15.2 & 81.7\\
Authority-aware & 68.0 & 55.7 & 60.9 & 87.2\\
Legacy conflict rollback & 64.6 & 85.2 & 22.2 & 86.5\\
Role-gated selector & 74.9 & 62.0 & 66.7 & 95.9\\
SIC+FrameMap & 86.0 & 62.0 & \textbf{100.0} & 95.9\\
\textbf{\method} & \textbf{95.1} & \textbf{85.2} & \textbf{100.0} & \textbf{100.0}\\
\bottomrule
\end{tabular}
\end{table}

\begin{table}[t]
\caption{MATRIX accuracy by question type. Each column contains 540 endpoint--item predictions.}
\label{tab:matrix-type}
\centering
\small
\begin{tabular}{lrrr}
\toprule
Method & Existence & Attribute & Relation\\
\midrule
Shifted CoT & 68.5 & 69.3 & 69.8\\
Legacy conflict rollback & 67.4 & 63.1 & 63.3\\
Role-gated selector & 72.4 & 75.7 & 76.5\\
SIC+FrameMap & 90.9 & 83.9 & 83.1\\
\textbf{\method} & \textbf{97.2} & \textbf{94.6} & \textbf{93.3}\\
\bottomrule
\end{tabular}
\end{table}

Table~\ref{tab:matrix-main} establishes \method as the strongest evaluated method on MATRIX. It reaches 95.1\%, exceeding SIC+FrameMap by 9.1 points and the role-gated selector by 20.2 points. The improvement is not obtained by simply favoring the image. Legacy conflict rollback matches \method at 85.2\% on misleading claims but collapses to 22.2\% on task rules, whereas \method preserves 100.0\% of task-defined answers. Simultaneously, clean-context accuracy rises to 100.0\%, showing that the hierarchy prevents both instruction overwriting and unnecessary correction.

The question-type breakdown in Table~\ref{tab:matrix-type} shows that the gain extends across existence, attribute, and relation reasoning. \method improves over SIC+FrameMap in every category and remains above 93\% even on relation questions. This breadth supports authority routing as a shared control mechanism over heterogeneous visual questions rather than a single-category heuristic.

\subsection{FRAME: Controlled Isolation of Route Precedence}
\begin{table}[t]
\caption{\framebench accuracy over 1,800 endpoint--item predictions.}
\label{tab:frame}
\centering
\small
\setlength{\tabcolsep}{4pt}
\begin{tabular}{lrrrr}
\toprule
Method & Overall & Mislead & Task & Clean\\
\midrule
Base CoT & 63.6 & 88.8 & 12.2 & 89.7\\
Visual-first & 75.5 & 51.0 & 76.0 & 99.5\\
Legacy conflict rollback & 68.3 & 90.5 & 15.7 & 98.7\\
Role-gated selector & 91.2 & 82.0 & 91.7 & 99.8\\
SIC & 99.1 & 97.5 & \textbf{100.0} & 99.8\\
SIC+FrameMap & \textbf{100.0} & \textbf{100.0} & \textbf{100.0} & \textbf{100.0}\\
\textbf{\method} & \textbf{100.0} & \textbf{100.0} & \textbf{100.0} & \textbf{100.0}\\
\bottomrule
\end{tabular}
\end{table}

\framebench disentangles precedence relations that are coupled in natural images. As Table~\ref{tab:frame} shows, \method reaches 100.0\% in all three regimes. Relative to SIC, frame recovery repairs 16 records without changing an already-correct prediction; task-first routing also prevents instruction overwrite under competing visual evidence.

\subsection{Paired Significance and Non-Regression}
\begin{table}[t]
\caption{Paired record-level analysis. CI denotes a record-cluster bootstrap interval; the sign test uses nonzero record differences.}
\label{tab:cluster}
\centering
\scriptsize
\setlength{\tabcolsep}{2.7pt}
\begin{tabular}{lrrrrr}
\toprule
Comparison & $\Delta$ (pp) & 95\% CI & Improved & Harmed & Sign-test $p$\\
\midrule
MATRIX: \method vs SIC+FrameMap & 9.07 & [6.79, 11.54] & 58 & 0 & $6.94\!\times\!10^{-18}$\\
MATRIX: \method vs role gate & 20.19 & [16.91, 23.58] & 126 & 0 & $2.35\!\times\!10^{-38}$\\
\framebench: \method vs SIC & 0.89 & [0.50, 1.33] & 16 & 0 & $3.05\!\times\!10^{-5}$\\
\framebench: \method vs SIC+FrameMap & 0.00 & [0.00, 0.00] & 0 & 0 & 1.0\\
\bottomrule
\end{tabular}
\end{table}

The paired analysis in Table~\ref{tab:cluster} confirms that the aggregate improvement is statistically strong and structurally targeted. On MATRIX, 58 of 270 records improve over SIC+FrameMap, none are harmed, and 212 tie. At endpoint level, \method changes 147 of 1,620 answers: 125 misleading-claim rows and 22 clean-context rows. Every changed answer is corrected, while all task-rule answers remain intact. The method is therefore a selective, non-regressive intervention rather than a global answer-bias shift.

\subsection{Endpoint Robustness and Route Consistency}
\begin{table}[t]
\caption{MATRIX overall accuracy by endpoint.}
\label{tab:provider}
\centering
\small
\setlength{\tabcolsep}{3.2pt}
\begin{tabular}{lrrrr}
\toprule
Endpoint & Shifted CoT & Authority & SIC+FrameMap & \method\\
\midrule
Gemma 3 & 77.0 & 74.4 & 83.0 & \textbf{94.8}\\
GPT-4o-mini & 68.1 & 70.0 & 88.1 & \textbf{96.3}\\
Llama 3.2 Vision & 60.0 & 59.6 & 88.1 & \textbf{90.7}\\
MiMo-V2-Omni & 67.0 & 53.3 & 87.0 & \textbf{96.7}\\
Mistral Small 3.2 & 70.0 & 68.1 & 84.1 & \textbf{95.2}\\
Qwen3-VL-Plus & 73.0 & 82.2 & 85.6 & \textbf{96.7}\\
\bottomrule
\end{tabular}
\end{table}

\method is strongest on every principal endpoint in Table~\ref{tab:provider}, improving over SIC+FrameMap by 2.6--11.8 points despite substantial variation in base model behavior. The route inventory also matches the executable policy: MATRIX contains 540 task-compilation actions, 540 context-to-text actions, 534 claim-to-anchor actions, and six fail-closed fallbacks; \framebench contains 600 task-compilation actions and 1,200 frame actions. These counts account for every prediction and show that one provider-agnostic hierarchy transfers across all evaluated endpoints without endpoint-specific thresholds or route tuning.

\subsection{Cross-Dataset Transfer on GQA-MATRIX}
\begin{table}[t]
\caption{GQA-MATRIX aggregate accuracy over 90 predictions from 30 records and three endpoints.}
\label{tab:gqa}
\centering
\small
\setlength{\tabcolsep}{5pt}
\begin{tabular}{lrrrr}
\toprule
Method & Overall & Mislead & Task & Clean\\
\midrule
Role-gated selector & 66.7 & -- & -- & --\\
SIC & 74.4 & 43.3 & 100.0 & 80.0\\
SIC+FrameMap & 74.4 & 43.3 & 100.0 & 80.0\\
\textbf{\method} & \textbf{93.3} & \textbf{80.0} & \textbf{100.0} & \textbf{100.0}\\
\bottomrule
\end{tabular}
\end{table}

On GQA-MATRIX, \method reaches 93.3\%, an 18.9-point gain over SIC+FrameMap. The largest improvement appears on misleading claims, where accuracy rises from 43.3\% to 80.0\%, while task preservation and clean-context non-interference remain at 100.0\%. All three endpoints obtain 28 correct answers out of 30. This result extends the observed routing behavior beyond the AMBER-derived construction, although the small cache supports a transfer check rather than a broad generalization claim.

\section{Limitations}
\method currently assumes a closed answer set and authority cues explicit enough for deterministic compilation or lexical resolution. MATRIX uses templated role realizations and \framebench uses standardized geometry; paraphrastic, implicit, adversarial, and open-ended authority remain important extensions. The replay protocol fixes candidate outputs to isolate routing behavior, so online candidate regeneration under provider drift is outside the present evaluation. Finally, the deterministic fallback returns a valid shared candidate rather than a calibrated abstention; extending the control layer with confidence-aware abstention is a natural next step.

\section{Conclusion}
We introduced mixed-authority vision-language reasoning and developed \method, an authority-informed control layer whose executable \sfr decides which source may govern a multimodal answer before correction occurs. By unifying task compilation, pixel-level frame recovery, semantic authority resolution, observable candidate routing, and fail-closed fallback, the method separates visual faithfulness from instruction obedience instead of forcing them into one modality preference. It reaches 95.1\% on MATRIX with a 9.1-point gain and zero harmed records, attains 100.0\% on \framebench, and transfers to 93.3\% on GQA-MATRIX. Across component additions, authority regimes, question types, endpoints, paired records, and controlled interventions, the evidence identifies explicit authority arbitration as the mechanism underlying robust selective correction.

\clearpage
\bibliographystyle{splncs04}
\bibliography{references}
\end{document}